\documentclass[letterpaper,10pt,conference]{ieeeconf}

\IEEEoverridecommandlockouts
\usepackage{cite}
\usepackage{amsmath,amssymb,amsfonts}
\usepackage{algorithmic}
\usepackage{graphicx}
\usepackage{textcomp}
\usepackage{xcolor}
\usepackage{subcaption}
\usepackage{float}
\usepackage{stfloats}
\usepackage[font=small]{caption}
\usepackage{hyperref}
\hypersetup{
    colorlinks=true,
    linkcolor=red,
    filecolor=magenta,      
    urlcolor=blue,
    pdftitle={darkgs},
    pdfpagemode=FullScreen,
    }
\usepackage{booktabs}
\usepackage{siunitx}
\usepackage{makecell}
\usepackage{xspace}
\usepackage{multirow}

\usepackage{stfloats}
\usepackage{colortbl}
\usepackage{pgf}
\usepackage{cleveref}

\def\BibTeX{{\rm B\kern-.05em{\sc i\kern-.025em b}\kern-.08em
    T\kern-.1667em\lower.7ex\hbox{E}\kern-.125emX}}
\begin{document}

\title{\LARGE Efficient Construction of Implicit Surface Models\\ From a Single Image for Motion Generation 
}

\author{
Wei-Teng Chu$^{*,1}$,
Tianyi Zhang$^{2}$,
Matthew Johnson-Roberson$^{5}$,
Weiming Zhi$^{3,4,5}$%
\thanks{$^{1}$Department of Electrical Engineering, Stanford University, USA.}
\thanks{$^{2}$Aurora Innovation, USA.}
\thanks{$^{3}$School of Computer Science, The University of Sydney, Australia.}
\thanks{$^{4}$Australian Centre for Robotics, The University of Sydney, Australia.}
\thanks{$^{5}$College of Connected Computing, Vanderbilt University, USA.}
\thanks{$^{*}$Correspondence to {\tt\small waynechu@stanford.edu}.}%
}

\maketitle

\begin{abstract}
Implicit representations have been widely applied in robotics for obstacle avoidance and path planning. In this paper, we explore the problem of constructing an implicit distance representation from a single image. Past methods for implicit surface reconstruction, such as \emph{NeuS} and its variants generally require a large set of multi-view images as input, and require long training times. In this work, we propose Fast Image-to-Neural Surface (FINS), a lightweight framework that can reconstruct high-fidelity surfaces and SDF fields based on a single or a small set of images. 
FINS integrates a multi-resolution hash grid encoder with lightweight geometry and color heads, making the training via an approximate second-order optimizer highly efficient and capable of converging within a few seconds. Additionally, we achieve the construction of a neural surface requiring only a single RGB image, by leveraging pre-trained foundation models to estimate the geometry inherent in the image. Our experiments demonstrate that under the same conditions, our method outperforms state-of-the-art baselines in both convergence speed and accuracy on surface reconstruction and SDF field estimation. Moreover, we demonstrate the applicability of FINS for robot surface following tasks and show its scalability to a variety of benchmark datasets. Code is publicly available at \url{https://github.com/waynechu1109/FINS}
\end{abstract}

\section{Introduction}

For autonomous robots to navigate and interact safely with the real world, they must form reliable geometric representations of their surroundings. Distance-based representations are a powerful representation widely used in motion planning and obstacle avoidance \cite{isdf, GeoFab_gloabL_opt, RAMP, CDF, driess2022learning, quintero2024stochastic, finean2021predicted, bukhari2025differentiable}. Accurate and efficient SDF estimation is therefore a key enabler of downstream decision-making and control.

Recent neural implicit surface methods, such as NeuS \cite{wang2021neus} and its successors \cite{neus2, volSDF, IDR, UNISurf, Gsurf}, have demonstrated impressive capability in reconstructing fine-grained object surfaces. However, these approaches suffer from two key drawbacks: (i) they rely on dense multi-view supervision, which is impractical in robotics settings where only sparse observations are available; and (ii) they require long training times, from minutes to hours, making them unsuitable for real-time use in navigation or manipulation. A complementary line of work \cite{GenS, long2022sparseneus, SparseCraft, SurfaceSplat} has sought to improve generalization to sparse views, reducing the dependency on extensive image collections. However, these approaches can often still require a sizeable number of images, and can be relatively inefficient to train from scratch.

In this paper, we introduce \emph{Fast Image-to-Neural Surface (FINS)}, a lightweight framework that overcomes these limitations. FINS reconstructs high-fidelity surfaces and SDF fields from as few as \emph{a single image}, or a small set of images, within seconds. Our framework integrates three components: (1) off-the-shelf 3D foundation models, such as DUSt3R \cite{dust3r_cvpr24} and VGGT \cite{wang2025vggt}, to lift single-view inputs into point clouds for SDF supervision; (2) a multi-resolution hash grid encoder \cite{instant-ngp} to enable efficient feature encoding; and (3) lightweight geometry and color heads trained with an approximate second-order optimizer, yielding rapid convergence.

\begin{figure}[t]
\centering
\includegraphics[width=0.499\linewidth]{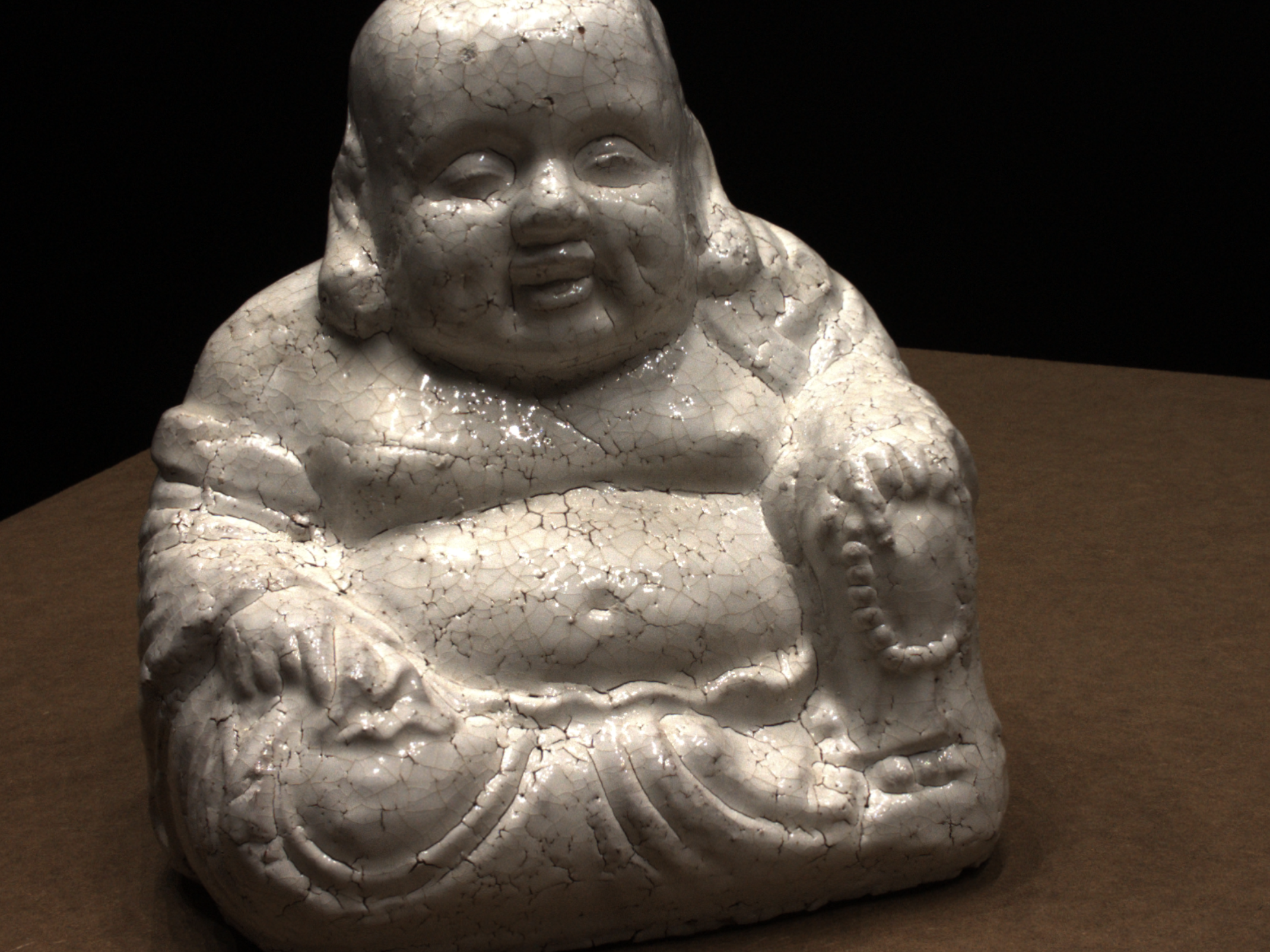}%
\includegraphics[width=0.499\linewidth]{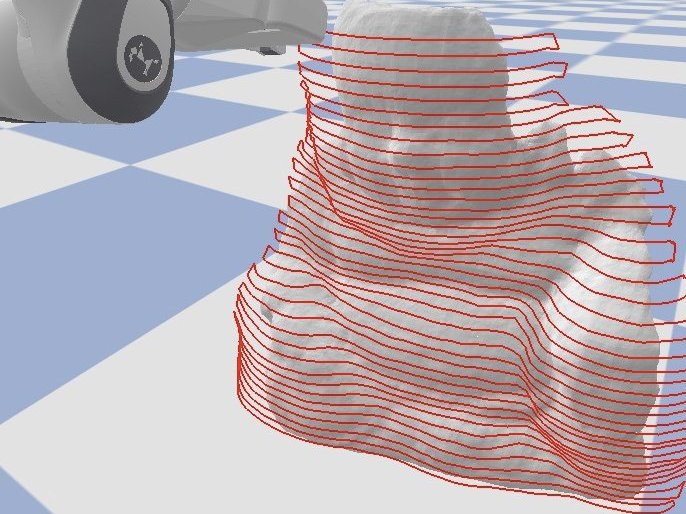}

\includegraphics[width=\linewidth]{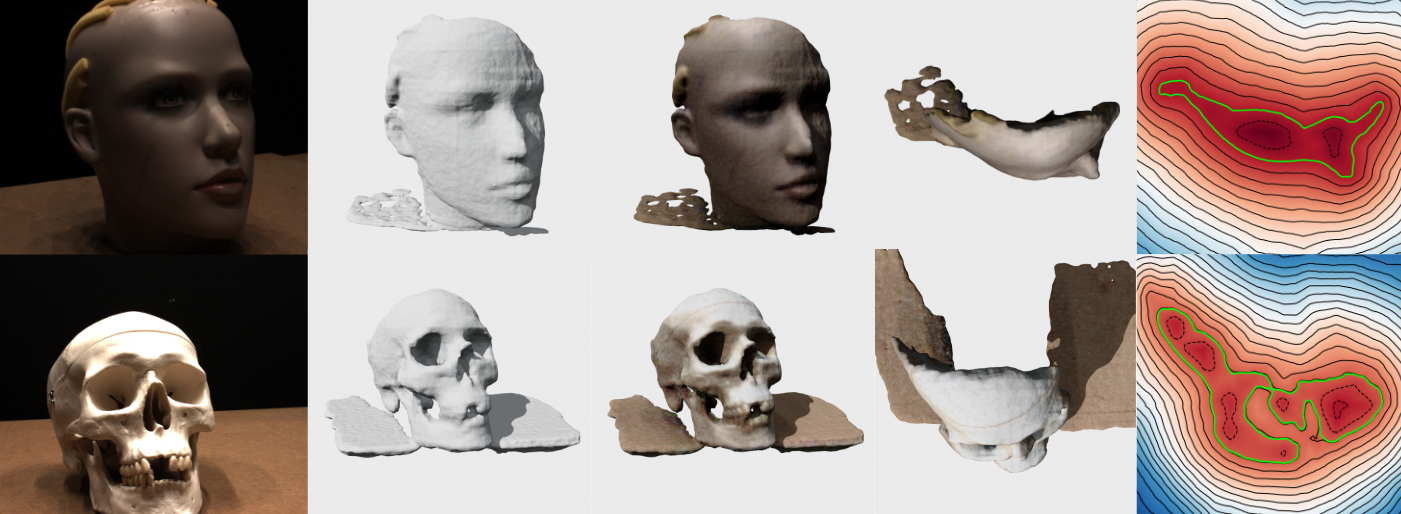}
\caption{We present Fast Image-to-Neural Surface (FINS), an efficient framework ($\sim$10s on consumer-grade hardware) that can reconstruct high-fidelity surfaces and SDF fields based on sparse or even a single image. \underline{Top row:} Input RGB image of a statue (left), and the corresponding implicit representation enabling robot motion to trace on the surface. \underline{Next two rows from left to right:} A single image input for SDF field reconstruction; The result mesh; The result colored mesh; The top view of the colored mesh; The trained SDF iso-contours corresponding to the top view.}
\label{intro_fig}
\end{figure}

By leveraging strong priors from pre-trained 3D models, FINS scales naturally from single objects to multi-view, scene-level settings. This flexibility supports deployment on mobile platforms, where continuous observations can be assimilated into an evolving SDF representation. As a result, FINS enables real-time reconstruction and refinement of neural surfaces for downstream robotics tasks such as obstacle avoidance, path planning, and surface following. We empirically evaluate the quality and efficiency of building implicit distance reconstructions of a diverse range of objects, and demonstrate the applicability of these representations for robot surface following \cite{periodic}. Concretely, this paper presents the following technical contributions:
\begin{enumerate}
    \item We propose FINS, an end-to-end method that achieves high-precision SDF training from a single image in only a few seconds.
    \item We leverage pre-trained 3D foundation models to generate point clouds for SDF supervision, enabling efficient and complete reconstruction with limited visual input.
    \item We adopt multi-resolution hash encoding and lightweight geometry/color heads with a mixed optimization strategy to eliminate heavy optimization and enable real-time convergence.
\end{enumerate}

\section{Related Work}

\textbf{Neural Implicit Surface Reconstruction:} Representations are critical in robotics \cite{Senanayke:2017, HM, vasudevan2025strategic, Vprism}. Neural implicit representations have become a dominant approach for 3D surface reconstruction. NeRF \cite{mildenhall2020nerf} pioneered neural radiance fields by learning volumetric scene representations with differentiable rendering. Subsequent works incorporated signed distance functions (SDFs) into the rendering pipeline, including VolSDF \cite{volSDF}, NeuS \cite{wang2021neus}, and NeuS2 \cite{neus2}, which enabled more consistent and detailed surface reconstruction. Extensions such as IDR \cite{IDR}, UNISurf \cite{UNISurf}, and GSurf \cite{Gsurf} further improved generalization and fidelity. Despite their success, these methods typically require dense multi-view supervision and lengthy optimization (tens of minutes to hours).

\textbf{Sparse-View and Generalizable Reconstruction:} To relax the dependence on dense viewpoints, several methods target sparse-view reconstruction. GenS \cite{GenS} enforced multi-scale feature-metric consistency to improve reconstruction under limited inputs. SparseNeuS \cite{long2022sparseneus} combined multi-level geometry reasoning with color blending and consistency-aware fine-tuning to enhance robustness to sparse views. SparseCraft \cite{SparseCraft} reconstructed detailed surfaces from very few inputs in under ten minutes, while SurfaceSplat \cite{SurfaceSplat} hybridized SDF-based and Gaussian splatting approaches for high-fidelity meshes. These methods significantly improve surface reconstruction from limited inputs. Other methods such as VGER \cite{VGER} looked at integrating video generators to push image sparsity. However, their focus remains on recovering meshes or surfaces, rather than constructing complete SDF fields, thereby limiting their utility for robotics tasks such as continuous collision checking and motion planning. 

\textbf{3D Foundation Models:} Large-scale 3D foundation models have recently shown remarkable progress in geometry estimation from sparse observations. DUSt3R \cite{dust3r_cvpr24} introduced a transformer-based architecture for dense correspondence and depth estimation, with subsequent improvements such as MONSt3R \cite{zhang2024monst3r}. An efficient model, VGGT \cite{wang2025vggt} learned generic 3D geometric priors from large-scale training. These models can generate high-quality point clouds from only a few images, providing strong geometric priors at low computational cost. 3D foundation models have been used within calibration \cite{JCR}, photorealistic reconstruction \cite{RecGS}, pose estimation \cite{Sim_pose}. However, these models do not directly yield SDF fields. Instead, they serve as powerful geometric initializations that can be integrated with implicit representations, making it feasible to construct accurate SDFs from limited views. This integration motivates our approach, where we leverage 3D foundation priors to enable real-time, single-image SDF reconstruction.

\section{Preliminaries}

\subsection{Multi-Resolution Hash Grid Encoding}
A fundamental problem in implicit neural field learning is how to embed spatial coordinates $\mathbf{x} \in \mathbb{R}^3$ into a high-dimensional representation that preserves geometric detail across scales. Standard sinusoidal positional encodings map coordinates to Fourier features, which expand the input into a fixed set of bases. While effective, this requires $\mathcal{O}(K)$ parameters per input dimension and leads to slow convergence in practice.

Instant-NGP \cite{instant-ngp} addresses this with a \emph{multi-resolution hash grid encoding}. Let $\{r_\ell\}_{\ell=1}^L$ denote a set of resolutions, where the $\ell$-th grid partitions space into $r_\ell^3$ cells. For a point $\mathbf{x}$, its grid coordinates at level $\ell$ are
\begin{align}
\mathbf{u}_\ell = r_\ell \cdot \mathbf{x}, \quad \mathbf{i}_\ell = \lfloor \mathbf{u}_\ell \rfloor, \quad \boldsymbol{\delta}_\ell = \mathbf{u}_\ell - \mathbf{i}_\ell,
\end{align}
where $\mathbf{i}_\ell \in \mathbb{Z}^3$ is the voxel index and $\boldsymbol{\delta}_\ell \in [0,1)^3$ is the local interpolation weight. Each grid vertex $\mathbf{v} \in \{\mathbf{i}_\ell + \mathbf{b} \mid \mathbf{b} \in \{0,1\}^3\}$ is mapped to an embedding $\mathbf{e}_\ell(\mathbf{v}) \in \mathbb{R}^F$ via a hash function
\begin{align}
\mathbf{e}_\ell(\mathbf{v}) = \Theta_\ell[h_\ell(\mathbf{v})], \quad h_\ell : \mathbb{Z}^3 \rightarrow \{1, \dots, T\},
\end{align}
where $\Theta_\ell \in \mathbb{R}^{T \times F}$ is a trainable hash table shared across the grid and $T \ll r_\ell^3$. The feature vector $\phi_\ell(\mathbf{x}) \in \mathbb{R}^F$ is computed by trilinear interpolation. Finally, the embeddings across $L$ levels are concatenated into a multi-scale representation
\begin{align}
E(\mathbf{x}) = \big[ \phi_1(\mathbf{x}), \phi_2(\mathbf{x}), \dots, \phi_L(\mathbf{x}) \big] \in \mathbb{R}^{L \cdot F}.
\end{align}

This scheme simultaneously encodes low-frequency structure (from coarse grids) and high-frequency detail (from fine grids) with constant memory $\mathcal{O}(L \cdot T \cdot F)$, independent of the native grid resolution. Empirically, this yields orders-of-magnitude faster convergence than Fourier features while maintaining compact parameterization.

\subsection{Approximate Second-Order Optimization}
Optimization dynamics significantly affect the quality of learned signed distance fields (SDFs). First-order optimizers such as SGD, AdamW, or Lion \cite{lion} update parameters $\theta$ using only gradient information:
\begin{align}
\theta_{t+1} = \theta_t - \eta \, \widehat{g}_t, \quad \widehat{g}_t \approx \nabla_\theta \mathcal{L}(\theta_t),
\end{align}
where $\mathcal{L}$ is the training loss and $\eta$ the learning rate. While cheap, such updates fail to account for the curvature of $\mathcal{L}$, leading to slow progress along high-curvature directions.  

Second-order methods instead rescale the update using the inverse Hessian matrix $H^{-1}$:
\begin{align}
\theta_{t+1} = \theta_t - \eta \, H^{-1} \nabla_\theta \mathcal{L}(\theta_t),
\end{align}
which preconditions the gradient according to the local geometry of the parameter space. However, computing and inverting $H$ exactly is intractable for large networks.

Kronecker-Factored Approximate Curvature (K-FAC) \cite{k-fac} factorizes each Fisher block layer-wise. For a fully connected layer with weights $W \in \mathbb{R}^{m \times n}$, inputs $a \in \mathbb{R}^n$, and pre-activation gradients $g \in \mathbb{R}^m$, the Hessian block is approximated as
\begin{align}
H_W \approx A \otimes G,\qquad & A = \mathbb{E}[a a^\top] \in \mathbb{R}^{n \times n}, \nonumber\\ & G = \mathbb{E}[g g^\top] \in \mathbb{R}^{m \times m}.
\end{align}
The Kronecker structure allows efficient inversion:
\begin{align}
H_W^{-1} \approx A^{-1} \otimes G^{-1}.
\end{align}

The natural gradient update for $W$ becomes
\begin{align}
\mathrm{vec}(\Delta W) = -\eta \, (A^{-1} \otimes G^{-1}) \, \mathrm{vec}(\nabla_W \mathcal{L}),
\end{align}
where $\mathrm{vec}(\cdot)$ denotes vectorization. This reduces the computational complexity from $\mathcal{O}((mn)^3)$ to $\mathcal{O}(m^3 + n^3)$ per layer, making curvature-aware optimization feasible at scale. In practice, K-FAC stabilizes training and accelerates convergence, serving as an effective middle ground between purely first-order and exact second-order methods in implicit field optimization.

\section{The Fast Image-to-Neural Surface Framework}
\subsection{Problem Formulation}
We address the task of learning a signed distance field (SDF) and corresponding appearance directly from sparse image observations. The input to our method is a single image or a small set of images, which are converted into 3D point clouds using off-the-shelf 3D foundation models. These point clouds serve as supervision for training the SDF.

Formally, each training input is a 3D coordinate
\begin{align}
\mathbf{x} = [x, y, z]^\top \in \mathbb{R}^3,
\end{align}
and the network maps this point to a 4D output vector containing the signed distance and RGB color values:
\begin{align}
f: \mathbb{R}^3 \to \mathbb{R}^4, \quad 
f(\mathbf{x}) = 
[d(\mathbf{x}), \, r(\mathbf{x}), \, g(\mathbf{x}), \, b(\mathbf{x})]^\top .
\end{align}

\subsection{Preprocessing with 3D Foundation Models}
Given an single RGB image or a small set of input images of an object or scene, we employ 3D foundation models such as DUSt3R \cite{dust3r_cvpr24} or VGGT \cite{wang2025vggt} to generate a colored point cloud. Each pixel in the input image is mapped to a 3D point with associated color and confidence, together with estimated camera intrinsics, extrinsics, and pose. The resulting point cloud therefore includes both object and background regions.  

\begin{figure}[t]
\centering
\includegraphics[width=0.5\linewidth]{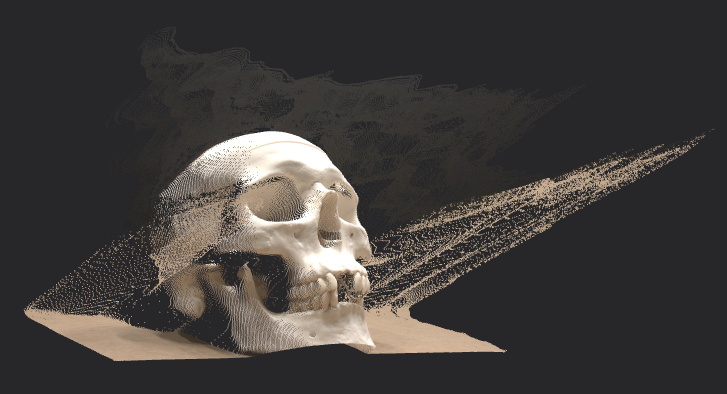}%
\includegraphics[width=0.48\linewidth]{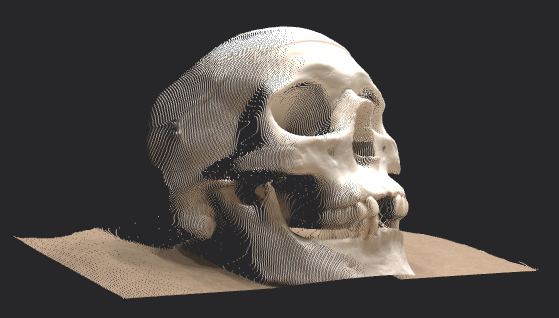}
\caption{we can leverage 3D foundation models to lift the skull image (shown in \cref{intro_fig}) to a 3D point cloud (left), then leverage confidence estimates to further filter and clean the point cloud (right).}
\end{figure}

To improve the quality of supervision, we filter out low-confidence points using the per-pixel confidence values predicted by the foundation model. This removes unreliable regions of the cloud while retaining dense and geometrically consistent supervision for SDF training.

\subsection{Model Design}
Our SDF network is an implicit neural representation consisting of a shared multi-resolution hash grid encoder and two prediction heads: a geometry branch (GeoNet) and a color branch (ColorNet). The encoder provides a compact, multi-scale embedding of the input coordinate, while the heads separately predict geometry and appearance.

\subsubsection{Multi-Resolution Hash Grid Encoding}
To efficiently capture both coarse and fine geometric details, we employ the multi-resolution hash grid encoding proposed by Instant-NGP \cite{instant-ngp}, which has since become standard in neural implicit surface modeling \cite{neus2, long2022sparseneus}. Our implementation uses $L = 10$ resolution levels, each producing a feature vector of dimension $F = 4$. The hash table size is $T = 2^{16}$ entries, with base resolution $R_{\text{base}} = 14$ and a per-level scaling factor of $s = 1.5$. This design allows the encoder to represent both low-frequency structure and high-frequency detail with a compact parameter budget.

\subsubsection{Geometry \& Color Heads}
The concatenated features from the hash encoder are processed by two lightweight branches:
\begin{itemize}
    \item \textbf{Geometry branch (GeoNet):} a two-layer MLP with Softplus activations that outputs the predicted signed distance $d(\mathbf{x})$.
    \item \textbf{Color branch (ColorNet):} a single linear layer that outputs the predicted RGB color vector $[r(\mathbf{x}), g(\mathbf{x}), b(\mathbf{x})]^\top$.
\end{itemize}
Separating geometry and appearance has been noted in \cite{wang2021neus,neus2} to improve training stability.

\subsection{Optimization Strategy}
A key novelty in FINS is an efficient \emph{staged hybrid optimization scheme}. We observe that the parameter sizes of the geometry and color networks can be sufficiently small to be trained via approximate second-order optimization. Here, we devise: 
\begin{itemize}
    \item \textbf{Warm-up stage (first 60\% of epochs):} all parameters are trained end-to-end with a standard first order optimizer. Here, we use the momentum-based first-order method, the Lion optimizer \cite{lion}.
    \item \textbf{Rapid Convergence (final 40\% of epochs):} the shared encoder continues to be updated using Lion, while the geometry and color heads are optimized using K-FAC \cite{k-fac}, a Kronecker-factored approximation to second-order optimization. This enables curvature-aware updates for the prediction heads while keeping encoder updates efficient.
\end{itemize}

This staged strategy balances rapid early learning with stable late-stage convergence, yielding both faster training and higher reconstruction accuracy.

\subsection{Training Objectives}
To jointly recover geometry and appearance, we adopt a composite multi-objective loss. The SDF must satisfy both local geometric fidelity (accurate surfaces) and global signed-distance consistency, while also matching photometric observations. A single objective is insufficient, so we combine complementary terms that enforce surface reconstruction, global regularization, and appearance supervision.

The full objective is expressed compactly as:
\begin{align}
\mathcal{L} 
= \sum_{t \in \mathcal{T}_{\text{surf}}} w_t \mathcal{L}_t
+ \sum_{t \in \mathcal{T}_{\text{reg}}} w_t \mathcal{L}_t
+ \sum_{t \in \mathcal{T}_{\text{rgb}}} w_t \mathcal{L}_t,
\end{align}
where $w_t$ are scalar weights. The index sets are
\begin{align}
\mathcal{T}_{\text{surf}} &= \{\text{SDF}, \text{zero}, \text{eik-surf}, \text{normal}\}, \nonumber\\
\mathcal{T}_{\text{reg}}  &= \{\text{eik-glob}, \text{sparse}, \text{off-surf}\},\nonumber \\
\mathcal{T}_{\text{rgb}}  &= \{\text{rgb}\}.\nonumber 
\end{align}
Each component loss term $\mathcal{L}_t$ plays a distinct role.

\textbf{SDF Loss}:  
    Supervises the predicted signed distance against ground-truth values at noisy points.
    \begin{equation}
    \mathcal{L}_{\text{SDF}} = \frac{1}{\sum_i w_i} \sum_{i \in \mathcal{N}} w_i \Big( d_\theta(x_i^{\text{noise}}) - d_i \Big)^2.
    \label{l_sdf}
    \end{equation}
    This ensures that the predicted SDF values faithfully reproduce the metric distances provided by supervision.

\textbf{Zero Loss}:  
    Encourages surface points to lie close to the zero-level set of the SDF.
    \begin{equation}
    \mathcal{L}_{\text{zero}} = \frac{1}{\sum_i w_i} \sum_{i \in \mathcal{N}} w_i \, \big| d_\theta(x_i) \big|.
    \label{l_zero}
    \end{equation}
    This helps anchor the reconstructed surface and prevents drift away from the observed boundary.

\textbf{Eikonal Loss}:  
    Enforces the Eikonal property $\|\nabla_x d(x)\|_2 = 1$ both near the surface and across the domain.
    \begin{align}
    \mathcal{L}_{\text{eik-surf}} &= \frac{1}{|\mathcal{M}|} \sum_{i \in \mathcal{M}} \Big(\|\nabla_x d_\theta(x_i)\|_2 - 1\Big)^2, \label{l_eik_surf} \\
    \mathcal{L}_{\text{eik-glob}} &= \frac{1}{|\mathcal{N}|} \sum_{i \in \mathcal{N}} \Big(\|\nabla_x d_\theta(\tilde{x}_i)\|_2 - 1\Big)^2, \label{l_eik_glob}
    \end{align}
    By enforcing unit-gradient norms, the network is guided toward a valid signed distance representation rather than an arbitrary scalar field.

\textbf{Normal Consistency Loss}:  
    Aligns predicted normals with ground-truth surface normals for stable geometry.
    \begin{equation}
    \mathcal{L}_{\text{normal}} = \frac{1}{|\mathcal{S}|} \sum_{i \in \mathcal{S}} \Big( 1 - \langle \hat{n}_i, n_i^{\text{gt}} \rangle \Big)^2,
    \label{l_normal}
    \end{equation}
    where $\hat{n}_i = \frac{\nabla_x d_\theta(x_i)}{\|\nabla_x d_\theta(x_i)\|}$. This sharpens surface quality and improves reconstruction of fine details.

\textbf{Sparse Regularization}:  
    Prevents the SDF from drifting by penalizing deviations from zero at randomly sampled points.
    \begin{equation}
    \mathcal{L}_{\text{sparse}} = \frac{1}{|\mathcal{N}|} \sum_{i \in \mathcal{N}} \exp\!\big(-\tau |d_\theta(\tilde{x}_i)|\big).
    \label{l_sparse}
    \end{equation}
    This discourages trivial solutions and improves stability when supervision is sparse.

\textbf{Off-Surface Loss}:  
    Enforces correct distances for explicitly sampled off-surface points.
    \begin{equation}
    \mathcal{L}_{\text{off-surf}} = \frac{1}{|\mathcal{N}_{\text{off}}|} \sum_{i \in \mathcal{N}_{\text{off}}} \Big( d_\theta(x_i) - d_i \Big)^2.
    \label{l_neg_sdf}
    \end{equation}
    This regularization ensures that the field encodes proper signed distance values even away from observed regions. Here, $\mathcal{N}_{\text{off}}$ is a set of sampled points off the SDF surface.

\textbf{RGB Reconstruction Loss}:  
    Supervises predicted colors against observed RGB values to ensure photometric consistency.
    \begin{equation}
    \mathcal{L}_{\text{rgb}} = 
    \frac{1}{|\mathcal{N}|} \sum_{i \in \mathcal{N}} 
    \big\| \mathbf{c}_\theta(x_i) - \mathbf{c}^{\text{gt}}_i \big\|_2^2.
    \label{l_rgb_gt}
    \end{equation}
    This couples geometry with appearance, ensuring that the reconstructed shape also reproduces surface-level visual cues.

This structured formulation ensures geometric fidelity at the surface, enforces global signed distance properties, and aligns appearance with observations, enabling stable convergence and high-quality reconstructions. We empirically validate the importance of each component loss term in \cref{subsec:ablation}.

\subsection{Surface Reconstruction}
Once trained, the SDF network can be used to recover explicit 3D geometry from implicit predictions. Each input is a 3D coordinate vector $\mathbf{x} \in \mathbb{R}^3$, corresponding to the spatial position of a sampled point in the reconstruction volume. To extract the geometry, we evaluate the network on a dense uniform grid of 3D samples within the reconstruction bounds, thereby constructing a volumetric SDF field representation of the scene. The iso-surface corresponding to $d(\mathbf{x}) = 0$ is then extracted using the marching cubes \cite{marching_cubes} algorithm, which produces a watertight triangle mesh. This mesh can also be colored by evaluating the model’s predicted RGB values at the corresponding grid points, yielding a textured reconstruction. 

For cleaner iso-surface extraction, we apply Gaussian smoothing to the raw SDF values in order to suppress high-frequency noise and improve mesh quality. Let $\mathbf{x}_i$ denote a grid point with predicted signed distance value $s_i$. The smoothed value $\tilde{s}_i$ is defined as:
\begin{equation}
\tilde{s}_i = 
\frac{\sum\limits_{j \in \mathcal{N}_G(\mathbf{x}_i)} 
\exp\!\left( -\frac{\|\mathbf{x}_i - \mathbf{x}_j\|^2}{2\sigma^2} \right) \cdot s_j}
{\sum\limits_{j \in \mathcal{N}_G(\mathbf{x}_i)} 
\exp\!\left( -\frac{\|\mathbf{x}_i - \mathbf{x}_j\|^2}{2\sigma^2} \right)},
\label{eq:gaussian_smoothing}
\end{equation}
where $\mathcal{N}_G(\mathbf{x}_i)$ is a local neighborhood around $\mathbf{x}_i$, and $\sigma$ is the smoothing bandwidth.  

This filtering process reduces artifacts from network prediction errors and enforces local smoothness in the reconstructed surface. By denoising the SDF prior to marching cubes, we obtain cleaner iso-surfaces with fewer spurious components and sharper, more visually coherent geometry.

\subsection{Robot Surface Tracing}\label{sec:methods_trace}
Implicit surfaces enable the construction of robot policies for \emph{surface tracing}. This is a class of commonly found problems, where a robot must follow the geometry of an object or environment at a certain distance. Examples include robotic inspection (e.g., crack detection or defect scanning), automated surface treatment (painting, polishing, or cleaning), and quality assurance tasks. In such settings, the robot’s end-effector is required first to approach a target distance from the surface, and then to move tangentially while maintaining contact or a fixed standoff distance. Here, we take a reactive motion generation approach \cite{Diff_templates} and model surface tracing with a piecewise (mode–switched) velocity field over the learned SDF $d(x)$, where $x\in\mathbb{R}^3$ is the end–effector, $k$ is controller gain, $d^\star$ is the desired iso–value, $\hat n(x) = \nabla d(x)/\|\nabla d(x)\|$ is the surface normal, and $P_T(x)=I-\hat n(x)\hat n(x)^\top$ projects onto the tangent plane. Using a small tolerance $\varepsilon>0$:
\begin{equation}
\dot{x} =
\begin{cases}
-\,k\,\big(d(x)-d^\star\big)\,\nabla d(x), 
& \text{if } |d(x)-d^\star| > \varepsilon \\
-\,k\,P_T(x)\,(x-x^\star), 
& \text{if } |d(x)-d^\star| \le \varepsilon.
\end{cases}
\end{equation}
The first mode exponentially drives the end–effector toward the desired iso–surface; once within the band $|d-d^\star|\le\varepsilon$, the controller switches to tangential motion that approaches the target $x^\star$ while remaining on the same contour (since $P_T$ removes the normal component). 

\begin{figure*}[t]
        \centering
            \includegraphics[width=0.9\linewidth]{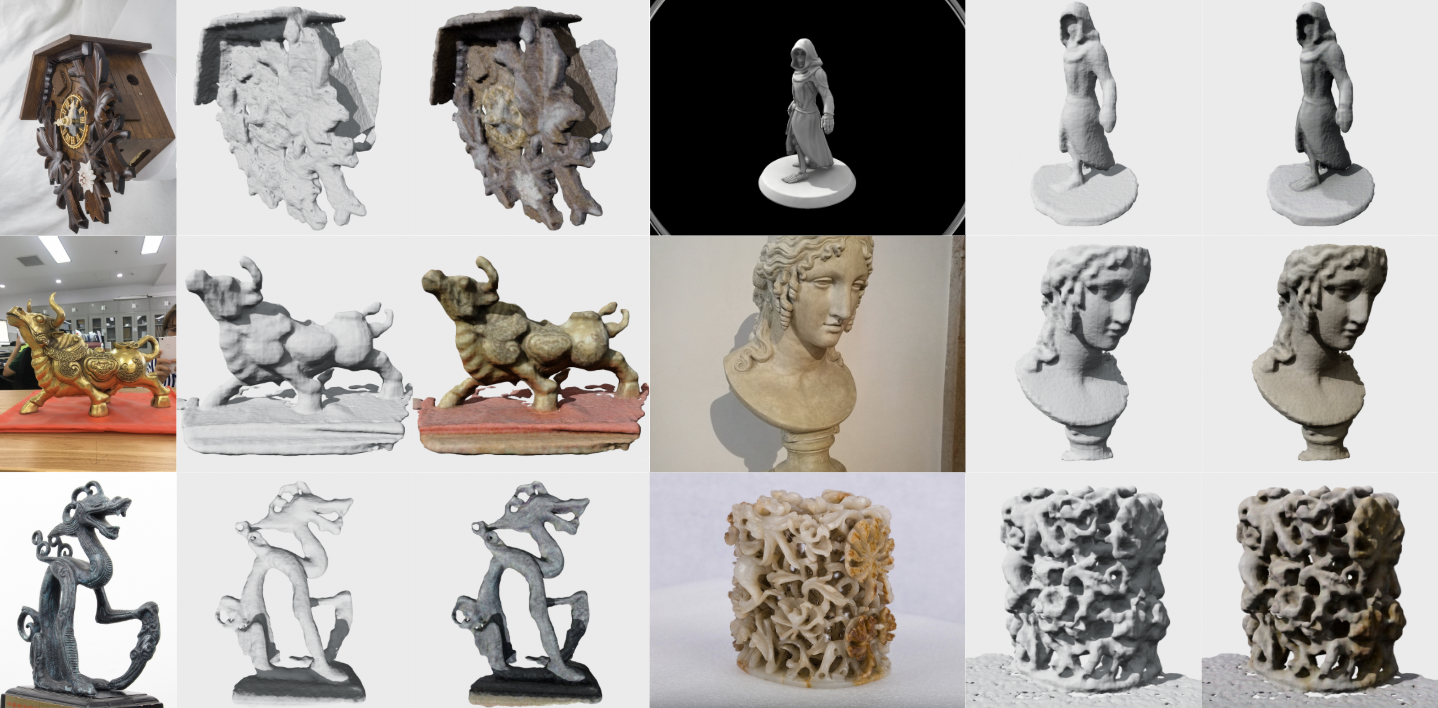}
                
        \includegraphics[width=0.9\linewidth]{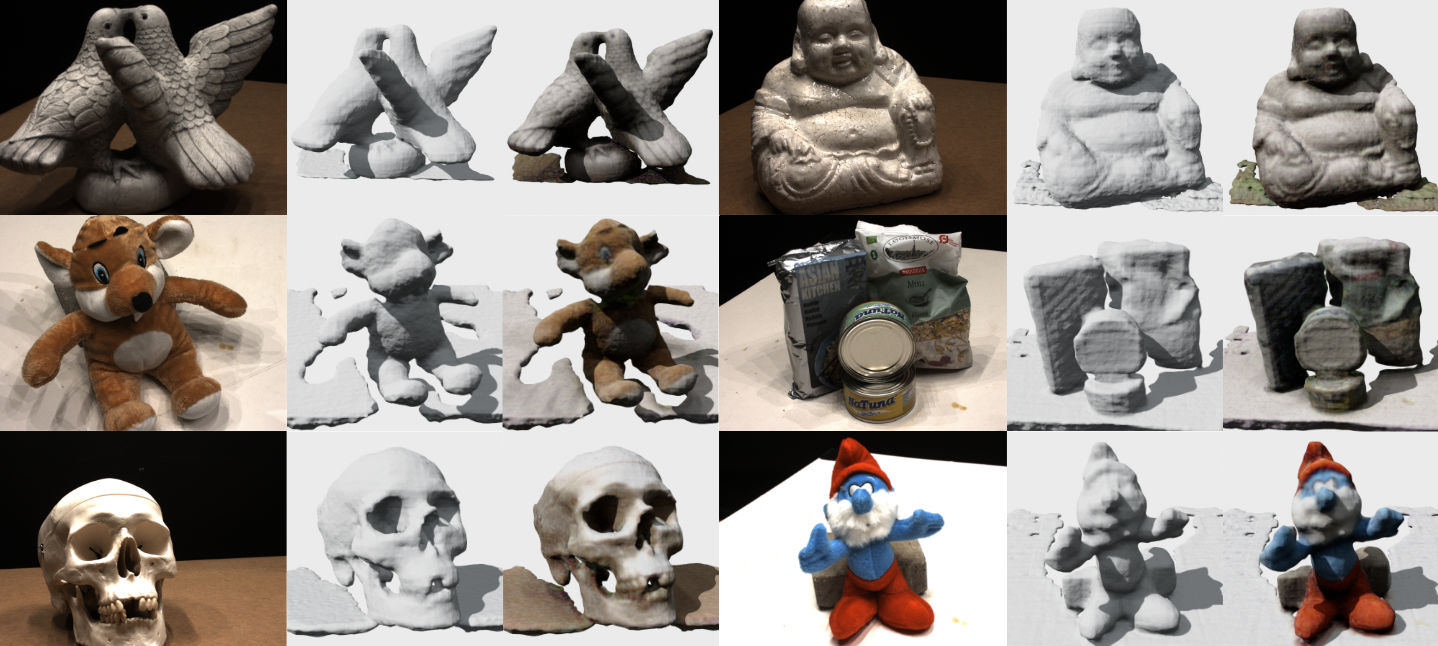}

    \caption{Qualitative reconstruction results on marching-cube visualized implicit surfaces generated from a \emph{single} input image from BlendedMVS \cite{blendedmvs} and DTU \cite{DTU} datasets. We illustrate the input image, the resulting geometry without color, and that with color.}
    \label{fig:more_results}
    \vspace{-2em}
\end{figure*}
\begin{table*}[t] 
\centering 
\resizebox{\textwidth}{!}{
\begin{tabular}{l c c | c c c c c c | c c c c} 
\toprule 
& & & \multicolumn{6}{c|}{DTU \cite{DTU}} & \multicolumn{4}{c}{BlendedMVS \cite{blendedmvs}} \\ 
\cmidrule(lr){4-9} \cmidrule(lr){10-13} 
Method & \makecell{\# of\\Input Images} & \makecell{Training\\Time (s)} 
& \multicolumn{3}{c}{CD ↓} & \multicolumn{3}{c|}{NAE (\si{\degree}) ↓} 
& \multicolumn{2}{c}{CD ↓} & \multicolumn{2}{c}{NAE (\si{\degree}) ↓} \\ 
\cmidrule(lr){4-6} \cmidrule(lr){7-9} \cmidrule(lr){10-11} \cmidrule(lr){12-13} 
& & & Smurf & Toy Tiger & Statue & Smurf & Toy Tiger & Statue & Sculpture & Bull & Sculpture & Bull \\ 
\midrule 
NeuS \cite{wang2021neus} & 49 & 247 & n/a & 11.83 & 8.07 & n/a & 8.58 & 10.83 & n/a & n/a & n/a & n/a \\ 
NeuS2 \cite{neus2} & 5 & 18 & 13.67 & \textbf{3.54} & \textbf{4.28} & \textbf{9.12} & 8.94 & 10.40 & 0.00890 & 0.0991 & \textbf{8.14} & 14.39 \\ 
SparseNeuS \cite{long2022sparseneus} & 2 & 127 & 16.10 & 5.57 & 8.18 & 9.90 & 9.01 & 10.39 & \textbf{0.0145} & 0.137 & 8.54 & 16.29 \\ 
SparseCraft \cite{SparseCraft} & 3 & 85 & 680.18 & 661.46 & 625.96 & 66.89 & 74.77 & 61.45 & 1.68 & 3.02 & 90.27 & 46.95 \\ 
\textbf{Ours} & \textbf{1} & \textbf{10} & \textbf{8.99} & 7.23 & 7.66 & 9.37 & \textbf{8.47} & \textbf{9.83} & 0.0198 & \textbf{0.0373} & 10.20 & \textbf{7.56} \\ 
\bottomrule 
\end{tabular} 
} 
\caption{Comparison of \textbf{Chamfer Distance ↓} and \textbf{Normal Angle Error ↓} across \textbf{DTU \cite{DTU}} and \textbf{BlendedMVS \cite{blendedmvs}} datasets.} 
\caption*{{{\footnotesize \textit{Note:} \textbf{n/a} denotes that the training did not converge. The experiments for SparseCraft are conducted on an NVIDIA A100 GPU due to the limited computational capability of the RTX 4060 Laptop GPU. The results for SparseNeuS are reported after 500 iterations of per-scene finetuning. }}}
\label{tab:diff_works_comparison} 
\vspace{-1em}
\end{table*}

\section{Experiments}
\subsection{Dataset and Metrics}
We evaluate the proposed FINS framework on examples from the DTU \cite{DTU} dataset and the BlendedMVS \cite{blendedmvs} dataset. The DTU dataset is one of the most widely used benchmarks in multi-view reconstruction, providing high-quality desktop scenes with accurate ground-truth annotations. In contrast, the BlendedMVS dataset is designed for large-scale multi-view stereo. Using these two datasets, we first compare our method against several baselines, focusing on the quality of the reconstructed meshes when trained for the same number of iterations. We then perform an ablation study on the loss terms in our model to examine the contribution of each design choice. Additionally, we examine the applicability of learned representations for robot surface-tracing tasks.  

All experiments are conducted on a single RTX 4060 Laptop GPU. For evaluation, we uniformly sample 200{,}000 vertices from both the reconstructed surface and the ground-truth surface. \textbf{Chamfer Distance (CD)} and \textbf{Normal Angle Error (NAE)} are employed as evaluation metrics to measure how closely the learned surface matches the ground truth. Specifically, given predicted surface points $P$ and ground-truth points $G$, the CD is defined as:
\begin{equation}
\text{CD}(P,G) = \frac{1}{|P|} \sum_{p \in P} \min_{g \in G} \|p-g\|_2^2 
+ \frac{1}{|G|} \sum_{g \in G} \min_{p \in P} \|g-p\|_2^2,
\end{equation}
and the NAE is defined as:
\begin{equation}
\text{NAE}(P,G) = \frac{1}{|P|} \sum_{p \in P} \min_{g \in G} 
\arccos \!\left( \frac{\langle n_p, n_g \rangle}{\|n_p\| \, \|n_g\|} \right),
\end{equation}
where $n_p$ and $n_g$ denote the normal vectors at points $p$ and $g$, respectively. The reported results are obtained by averaging over five independent runs. The specific data we use for quantitative evaluations are \emph{Statue}, \emph{Toy Tiger}, \emph{Smurf} (labelled in DTU as Scans 114, 105, and 82 respectively); \emph{Sculpture} and \emph{Bull} are selected from the BlendedMVS dataset.

\begin{table*}[t]
  \centering

  \begin{tabular}{l | c c c c c c | c c c c}
    \toprule
           & \multicolumn{6}{c|}{DTU \cite{DTU}} & \multicolumn{4}{c}{BlendedMVS \cite{blendedmvs}} \\
    \cmidrule(lr){2-7} \cmidrule(lr){8-11}
    Variant & \multicolumn{3}{c}{CD ↓} & \multicolumn{3}{c|}{NAE (\si{\degree}) ↓} & \multicolumn{2}{c}{CD ↓} & \multicolumn{2}{c}{NAE (\si{\degree}) ↓} \\
    \cmidrule(lr){2-4} \cmidrule(lr){5-7} \cmidrule(lr){8-9} \cmidrule(lr){10-11}
           & Smurf & Toy Tiger & Statue & Smurf & Toy Tiger & Statue & Sculpture & Bull & Sculpture & Bull \\
    \midrule
    \textbf{Full model}                & 8.99 & 7.23 & 7.66 & 9.37 & \textbf{8.47} & \textbf{9.83} & 0.0198 & 0.0373 & 10.20 & \textbf{7.56} \\
    w/o $\mathcal{L}_{\text{SDF}}$     & 10.37 & 7.92 & 7.81 & 9.96 & 9.16 & 10.02 & 0.0138 & 0.0439 & 9.85 & 9.27 \\
    w/o $\mathcal{L}_{\text{zero}}$    & 12.94 & 14.26 & 8.58 & 9.64 & 8.99 & 9.87 & 0.0238 & 0.0761 & 10.29 & 9.13 \\
    w/o $\mathcal{L}_{\text{eik}}$     & \textbf{5.77} & \textbf{5.94} & \textbf{4.67} & \textbf{9.02} & 8.72 & 10.22 & \textbf{0.0134} & 0.0400 & 10.90 & 8.75 \\
    w/o $\mathcal{L}_{\text{normal}}$  & 10.01 & 7.54 & 8.20 & 9.39 & 9.45 & 10.41 & 0.0192 & 0.0441 & 10.70 & 8.51 \\
    w/o $\mathcal{L}_{\text{sparse}}$  & 9.02 & 7.14 & 6.62 & 9.38 & 8.86 & 10.04 & 0.0209 & 0.0378 & 10.24 & 7.62 \\
    w/o $\mathcal{L}_{\text{off-surf}}$ & 8.22 & 6.56 & 5.83 & 9.32 & 9.42 & 10.14 & 0.0167 & \textbf{0.0368} & \textbf{8.82} & 7.80 \\
    \bottomrule
  \end{tabular}
  \caption{Ablation study of different design choices in our method on \textbf{DTU \cite{DTU}} and \textbf{BlendedMVS \cite{blendedmvs}} datasets, reported on multiple scans with \textbf{Chamfer Distance ↓} and \textbf{Normal Angle Error ↓}.}
  \label{tab:ablation}
  \vspace{-2em}
\end{table*}

\subsection{Evaluation Details}
For examples from the DTU dataset, we adopt the reference meshes provided by the benchmark, specifically the Poisson surface reconstructions of the MVS point clouds generated by Tola et al.~\cite{tola2012efficient}. For the BlendedMVS dataset, we directly use the ground-truth meshes released by the authors.  

\textbf{Mesh Alignment.}  
To ensure fair and consistent evaluation, all reconstructed meshes are aligned to the coordinate frame of the ground truth prior to metric computation. Without alignment, even small pose discrepancies can dominate Chamfer Distance and Normal Angle Error, obscuring the effect of reconstruction quality. Prior to alignment, large planar structures (e.g., desktops) that are present in both predicted and ground-truth meshes are manually removed using MeshLab, as they otherwise bias the correspondence search. Alignment is then performed in two stages:  

\emph{Rough alignment:} several pairs of corresponding landmarks are manually annotated between the reconstructed mesh and the ground truth. These correspondences provide a set of 3D point matches, which are used to compute an initial similarity transform with the Umeyama method~\cite{Umeyama}. This step resolves large-scale differences in orientation, translation, and scale.  

\emph{Fine alignment:} to refine the initial transform, both meshes are uniformly sampled to generate dense point clouds. These point clouds are then aligned using point-to-point Iterative Closest Point (ICP)~\cite{icp}, which minimizes residual misalignment and brings the meshes into near-perfect correspondence.  

Since the rough alignment depends on manually selected correspondences, perfect alignment cannot be guaranteed, and slight residual errors may affect the reported scores. Nevertheless, for each object case the same alignment transform is applied consistently across all methods, ensuring that the comparison remains fair and unbiased.

\subsection{Object Surface Reconstruction}
We next compare our approach against several baselines on object-level surface reconstruction. For consistency, we train all methods for 500 epochs.  
As summarized in Table~\ref{tab:diff_works_comparison}, \textsc{FINS} achieves competitive or superior reconstruction quality across many of the evaluated objects in the DTU and BlendedMVS datasets, while requiring dramatically fewer resources in terms of both input images and training time.  

NeuS~\cite{wang2021neus} relies on dense supervision with $49$ input images and more than $240$ seconds of optimization per scene. Even with this heavy supervision, its performance is inconsistent and in some cases fails to converge, making it impractical for robotic scenarios where rapid scene understanding is needed. NeuS2~\cite{neus2} improves efficiency substantially, requiring only $5$ input views and $18$ seconds of training. It achieves the lowest Chamfer Distance on DTU’s \emph{Toy Tiger} ($3.54$) and \emph{Statue} ($4.28$) objects, demonstrating strong accuracy. However, the requirement of multiple input views remains restrictive when only a single opportunistic observation is available in robotic deployment. SparseNeuS~\cite{long2022sparseneus} further reduces the number of required inputs to $2$ images, but finetuning times remain over $120$ seconds per scene, and while its Normal Angle Error is competitive (e.g., $8.54$\si{\degree} on \emph{Sculpture}), its Chamfer Distance is significantly higher (e.g., $16.10$ on \emph{Smurf}). SparseCraft~\cite{SparseCraft}, despite leveraging an NVIDIA A100 GPU, produces divergent results: Chamfer Distances in the hundreds (e.g., $680.18$ on \emph{Smurf}) and Normal Angle Errors exceeding $45$\si{\degree}, coupled with training times beyond $80$ seconds. 

In contrast, \textsc{FINS} reconstructs detailed geometry and a consistent signed distance field from only a \emph{single RGB image}, converging in approximately $10$ seconds on a consumer-grade RTX 4060 Laptop GPU. On DTU, our method achieves Chamfer Distances of $8.99$ (\emph{Smurf}), $7.23$ (\emph{Toy Tiger}), and $7.66$ (\emph{Statue}), along with Normal Angle Errors consistently around $7$\si{\degree}–$10$\si{\degree}. On BlendedMVS, \textsc{FINS} attains strong results across both indoor and outdoor categories, with Normal Angle Errors as low as $7.56$ (\emph{Bull}), competitive with or better than all baselines. Although \textsc{FINS} does not always outperform NeuS2 on every metric, it consistently balances accuracy with extreme efficiency, reducing both input views (from $5$–$49$ views to just $1$) and training time (from $18$–$600+$ seconds to only $10$).

\begin{figure*}[t]
\centering
    \includegraphics[width=0.49\linewidth]{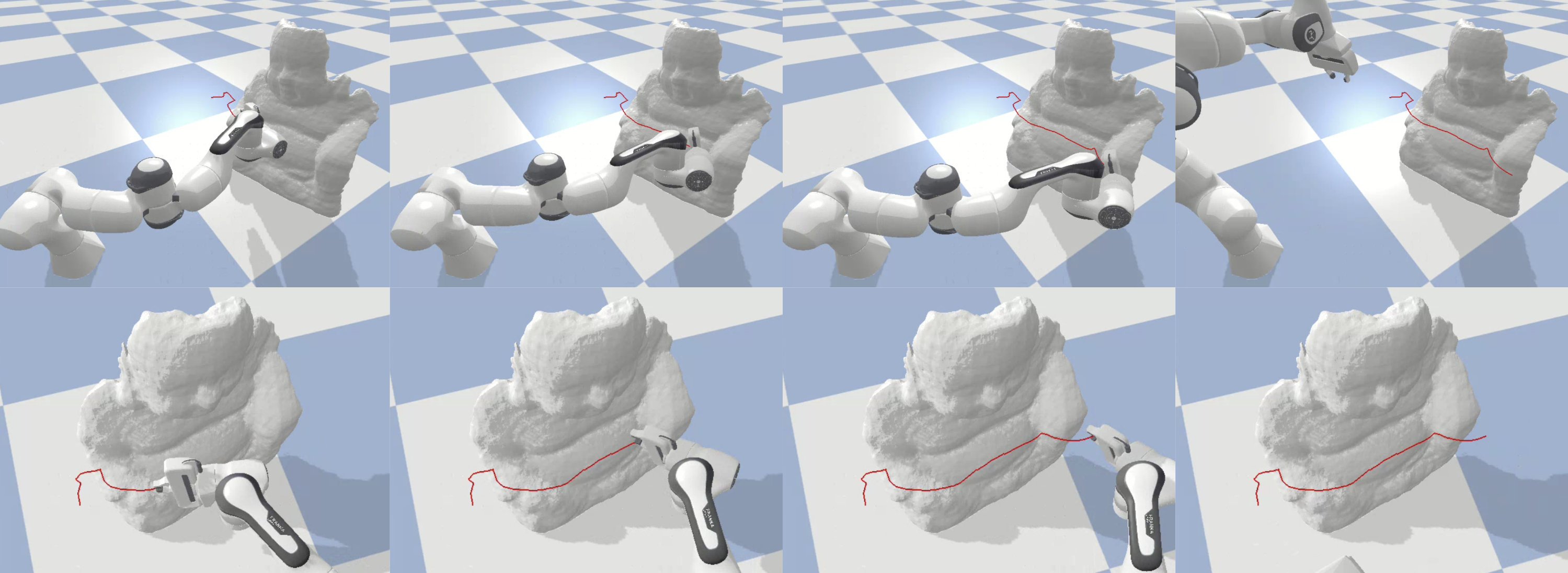}
    \includegraphics[width=0.49\linewidth]{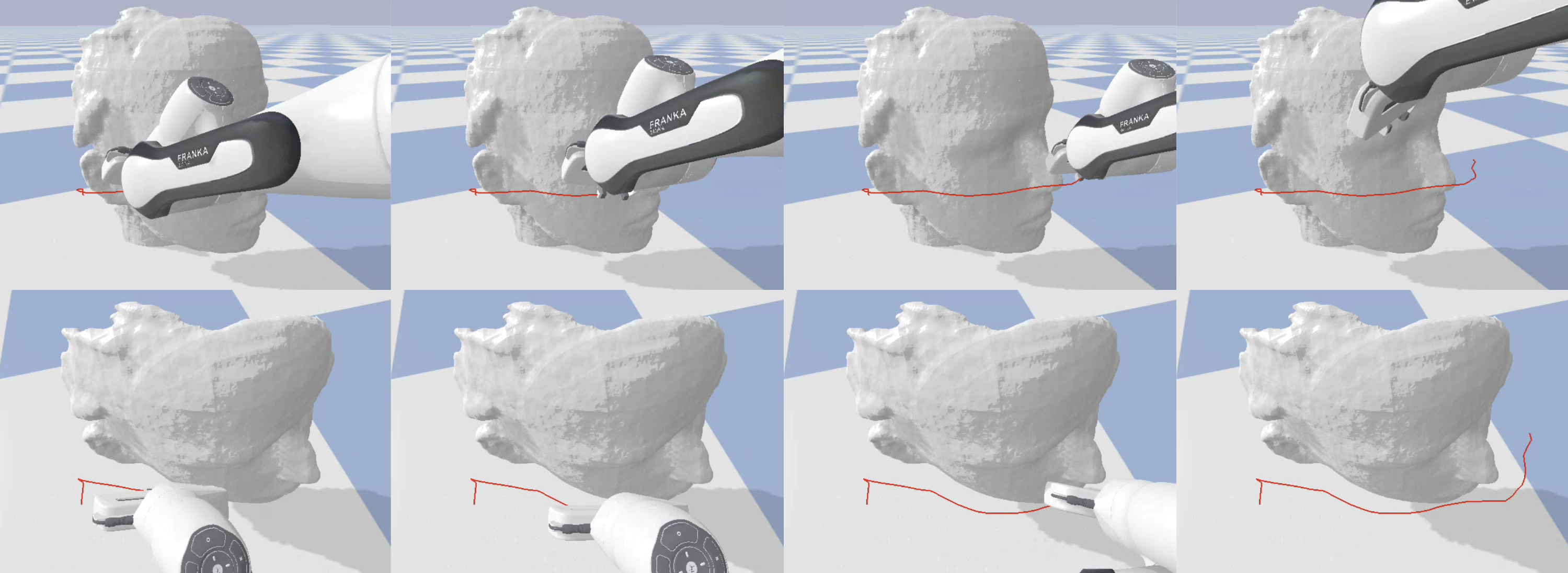}
    \caption{The implicit distance representation produces accurate iso-surfaces, which enable robot surface-tracing motion generation. The robot's motion can be generated by considering the normal and gradient vectors of iso-surfaces of the learned model, tracing the surface of reconstructions of the \emph{Statue} and \emph{Head} images. The red line denotes the Franka's end effector path.}
    \label{fig:robot_sim}
    \vspace{-1em}
\end{figure*}

\begin{table}[t]
  \centering
  \resizebox{\linewidth}{!}{%
    \begin{tabular}{l c | c c c c}
      \toprule
      \multicolumn{1}{c}{Variant} & Time (s) & \multicolumn{2}{c}{CD ↓} & \multicolumn{2}{c}{NAE (\si{\degree}) ↓} \\
      \cmidrule(lr){3-4} \cmidrule(lr){5-6}
                                  &          & Toy Tiger & Statue & Toy Tiger & Statue \\
      \midrule
      HE + Mixed 2nd Order & \textbf{10} & \textbf{7.23} & 7.66 & \textbf{8.47} & \textbf{9.83} \\
      PE + 1st Order   & 219 & 7.89 & \textbf{7.57} & 9.14 & 10.55 \\
      \bottomrule
    \end{tabular}%
  }
  \caption{Comparison between \textbf{Hash Encoding w/ 2nd-order optimizer (Ours)} and \textbf{Positional Encoding w/ 1st-order optimizer} on the DTU dataset, evaluated using \textbf{Chamfer Distance ↓} and \textbf{Normal Angle Error ↓}.}
  \label{tab:ablation_hepe}
\end{table}

\begin{figure}[t]
    \centerline{\includegraphics[width=0.9\linewidth]{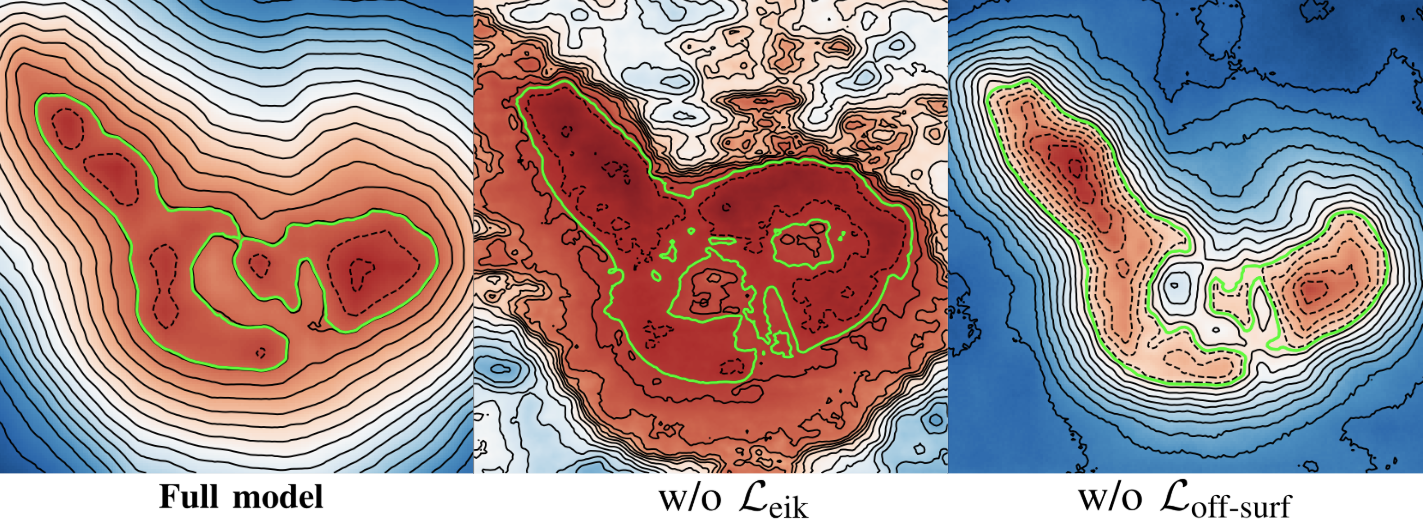}}
    \caption{Removing the Eikonal or Off-surface loss term can lead to a better surface reconstruction quality, which can lead to poor contours off the surface of the representation.}
    \label{fig: ablation_compare}
    \vspace{-1.5em}
\end{figure}

\subsection{Ablation Study}\label{subsec:ablation}
We analyze the contribution of each loss and our optimization/encoding choices using single–image reconstruction on DTU and BlendedMVS. All variants are trained for 500 iterations with identical data, sampling, and schedules.

\textbf{Effect of loss terms:}
Table~\ref{tab:ablation} reports Chamfer Distance (CD) and Normal Angle Error (NAE) for the full model and for variants with individual losses removed. The \emph{full} configuration delivers consistently strong geometry across objects and datasets, indicating that the objectives are complementary. Interestingly, removing the Eikonal regularizer can sometimes reduce CD on DTU (e.g., \emph{Smurf}/\emph{Toy Tiger}), but overfits. This is shown in Fig.~\ref{fig: ablation_compare} where this overfitting comes at the expense of a valid signed–distance structure: gradients lose unit norm away from the surface, producing distorted level sets and degraded SDF quality. Similar trends appear when dropping the zero-level constraint or normal consistency: surfaces may remain visually plausible, yet normals become noisy and level sets drift, which is detrimental for downstream planning. Overall, enforcing (i) zero-crossing consistency, (ii) unit-norm gradients, and (iii) normal alignment stabilizes the field and preserves geometry in low-supervision regimes.

\textbf{Encoding and optimizer:}
We further compare our multi-resolution hash encoding with mixed first/second-order training against a conventional positional encoding with a first-order optimizer (Table~\ref{tab:ablation_hepe}). The hash+K-FAC variant attains comparable or better accuracy while reducing wall-clock time from minutes to \(\sim\)10\,s on a laptop GPU. The combination yields fast early progress via Lion on the shared encoder and curvature-aware refinement of the small geometry/color heads via K-FAC, improving convergence stability without incurring the cost of full second-order updates on the entire network.

\textbf{Takeaways:}
The various loss terms, including the Eikonal, zero-level, and normal terms are jointly necessary to obtain high-quality SDFs, even when raw mesh metrics (CD) which measure surface quality momentarily improve without them. Hash encoding plus lightweight, head-only second-order updates offers a favorable accuracy–time trade-off, enabling practical single-image reconstruction efficiently.

\begin{figure}[t]
\centering
\includegraphics[width=0.33\linewidth]{figures/buddha_in.png}%
\includegraphics[width=0.33\linewidth]{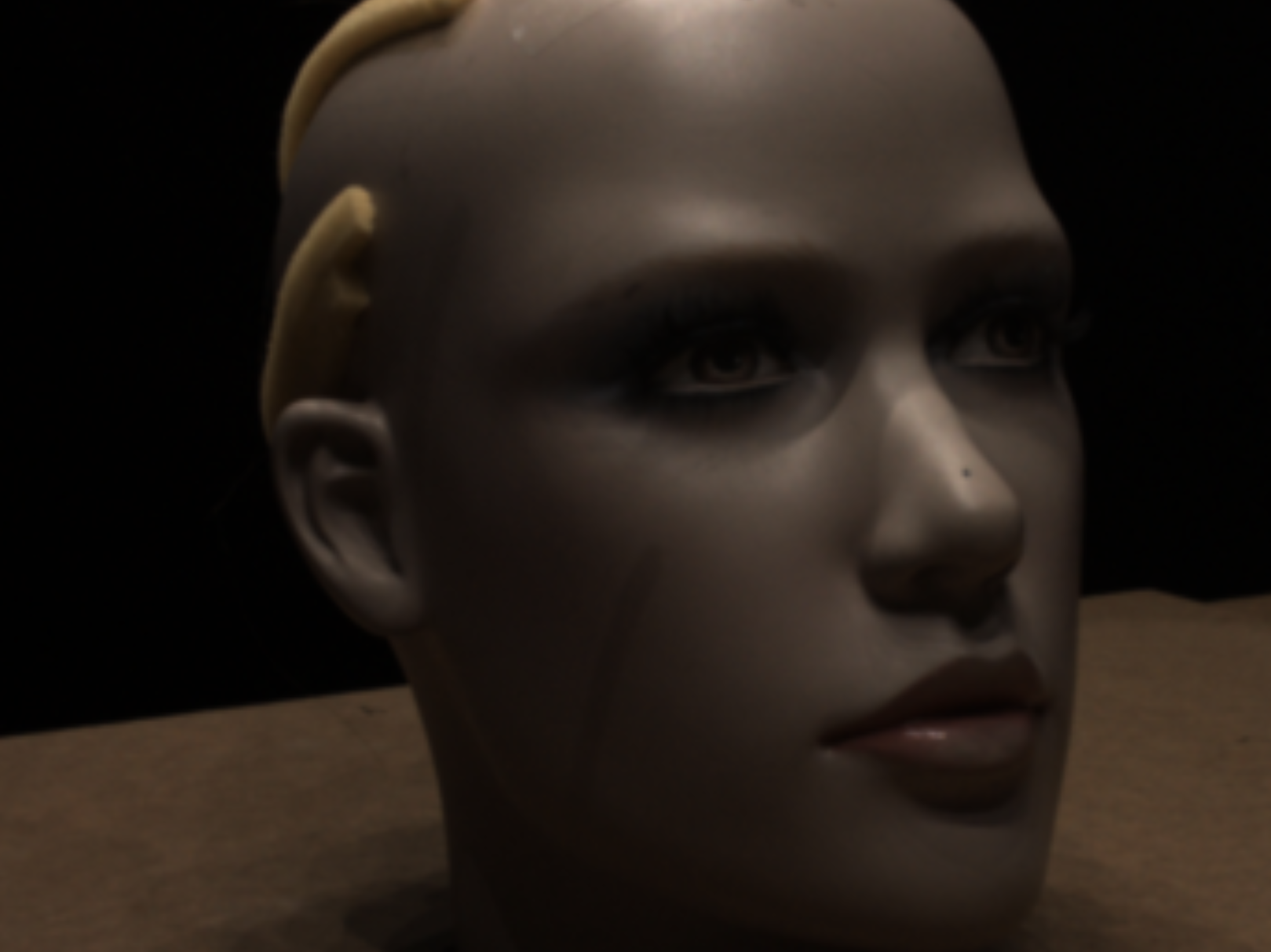}%
\includegraphics[width=0.33\linewidth]{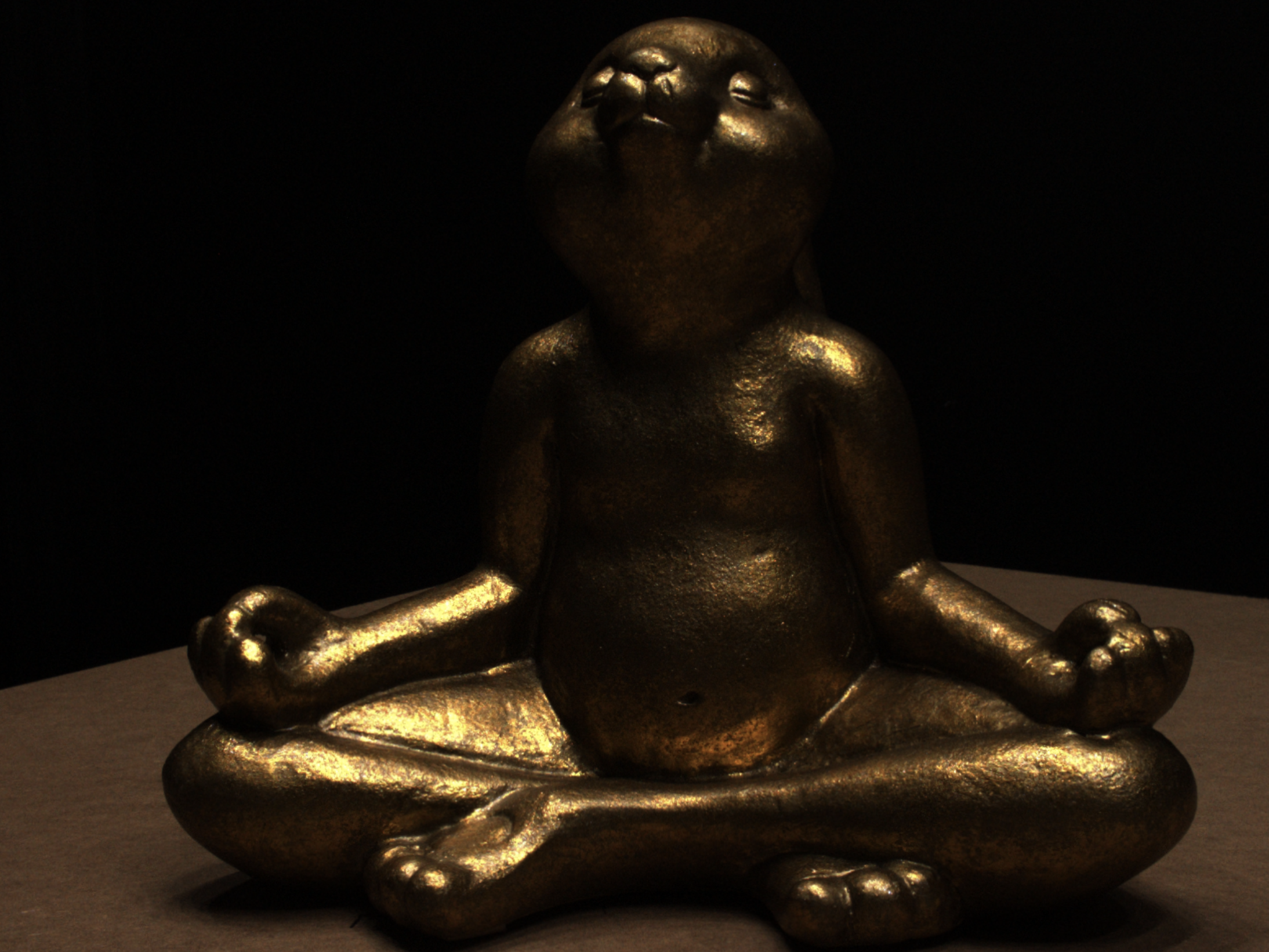}%

\includegraphics[width=0.33\linewidth]{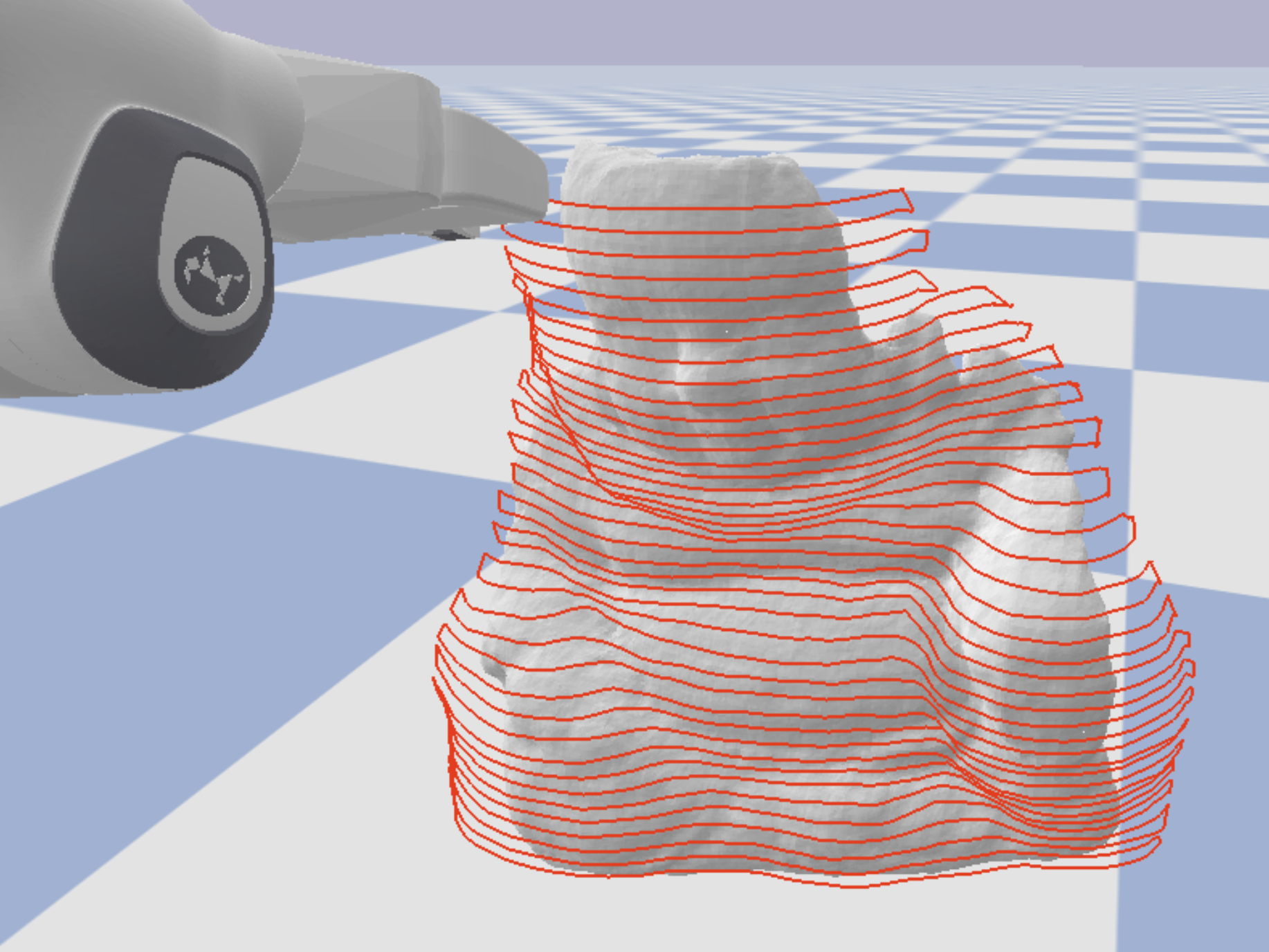}%
\includegraphics[width=0.33\linewidth]{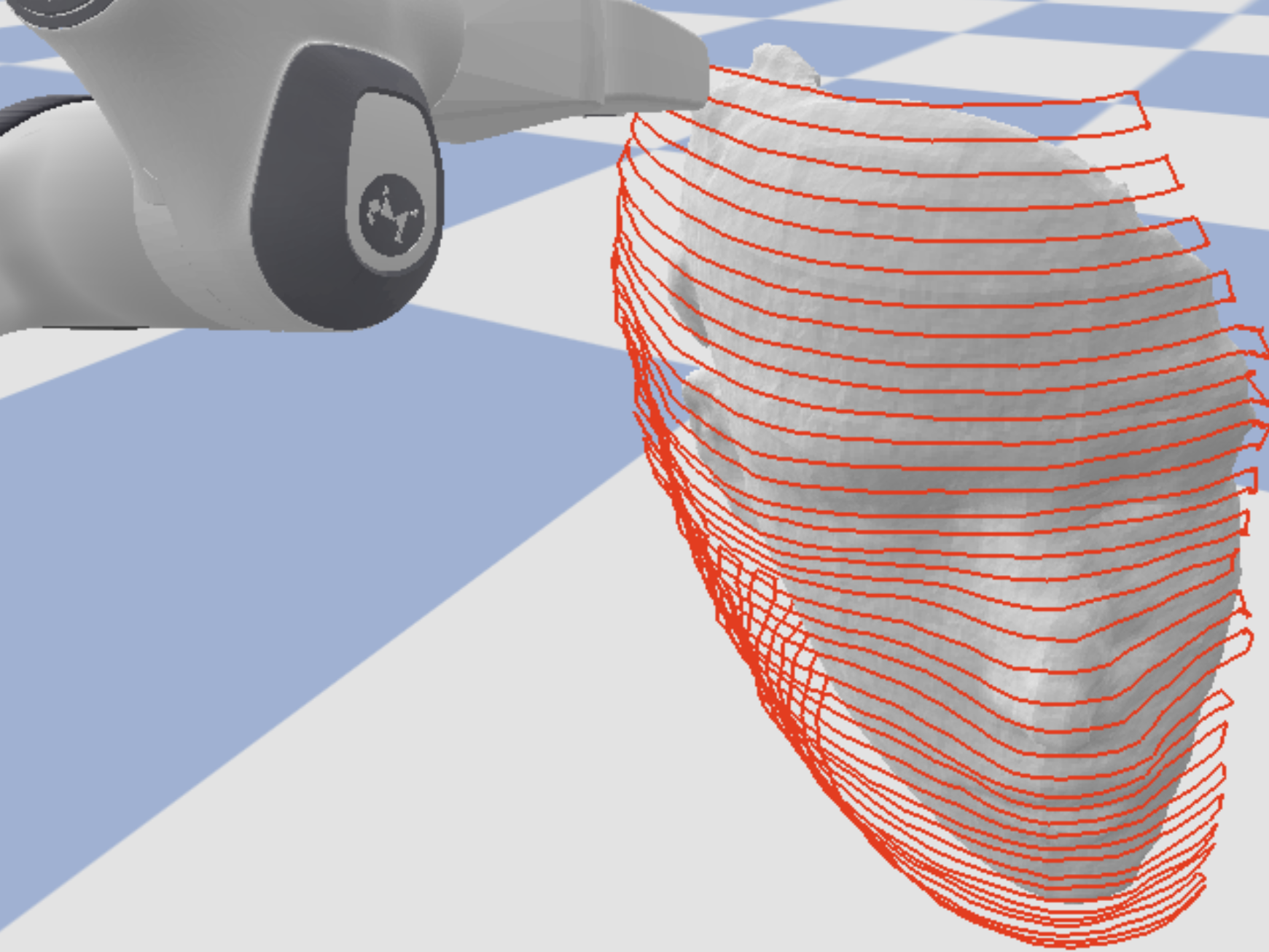}%
\includegraphics[width=0.33\linewidth]{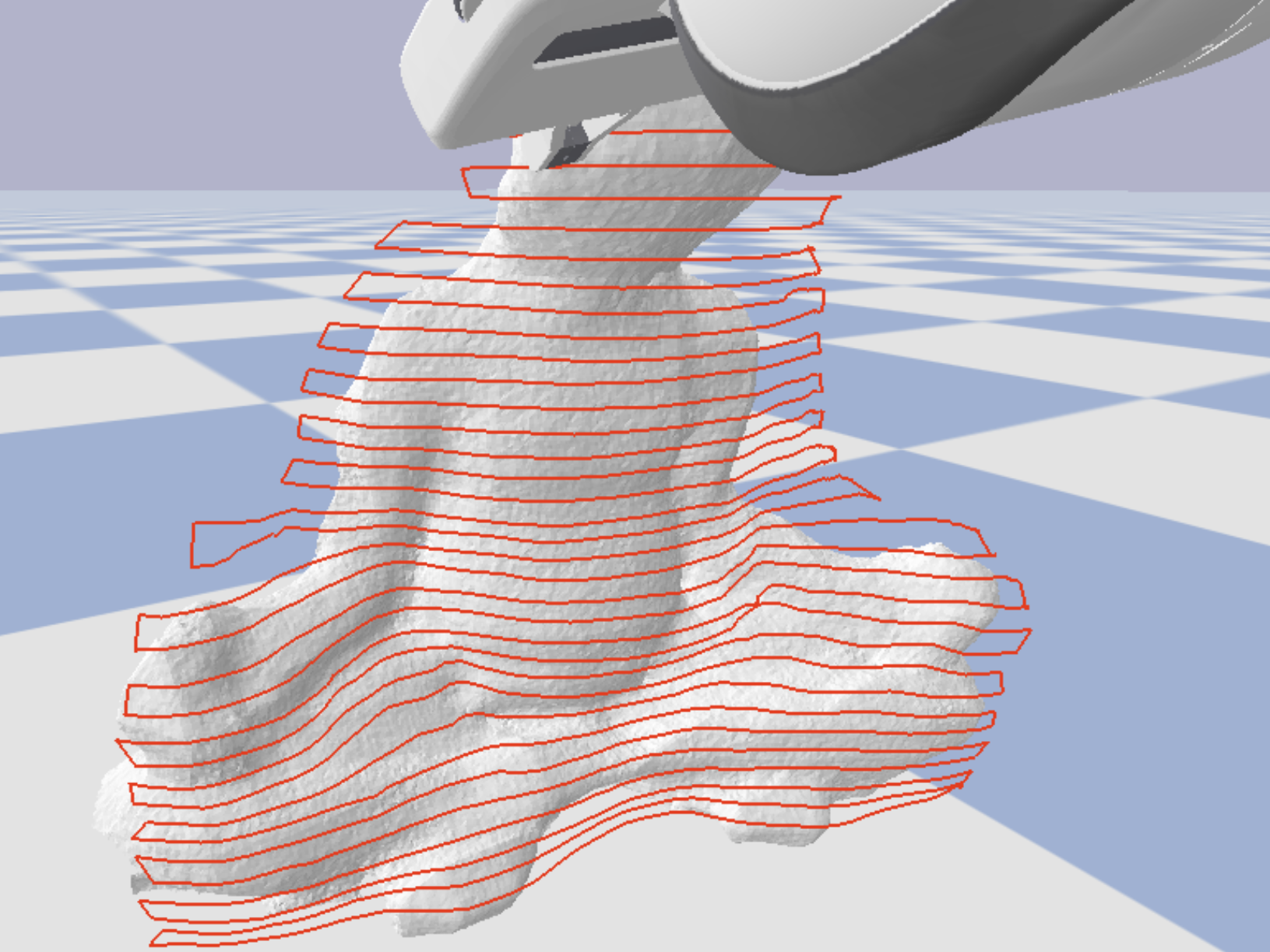}%
\caption{FINS takes in single images (top), and produces implicit representations, which allow surface following. Lawn mower patterns tracing off the surface are shown in red (bottom).}\label{fig:lawn}
\end{figure}

\subsection{Robot Surface Tracing}
We validate the practical utility of the learned SDFs by applying them to a surface–tracing task in the PyBullet simulator \cite{pybullet} with a Franka Emika Panda arm. This task requires the end–effector to approach a desired standoff distance from the reconstructed surface and then follow the surface tangentially, as commonly needed in inspection, painting, or polishing.

The controller implemented is outlined in Section~\ref{sec:methods_trace}, using a piecewise velocity field defined over the learned SDF $d(x)$. In the \emph{approach phase}, the end–effector is driven toward the target iso–value $d^\star$ by following the SDF gradient, ensuring exponential convergence to the desired contour. Once within a tolerance band, the controller switches to the \emph{surface–following phase}, in which tangential motion is generated by projecting the velocity onto the local tangent plane. This guarantees that motion remains constrained to the iso–surface while progressing toward the goal. In simulation, we set $d^\star=0.05$ and commanded the robot to trace along the reconstructed meshes. Figure~\ref{fig:robot_sim} shows the Panda arm a the reconstructed object, by tracing hugging the surface at the specified offset, by following the iso-surface of our learned model. Additional examples of running lawn-mower patterns by considering the iso-surfaces of the learned implicit representations are shown in \cref{fig:lawn}. These experiments confirm that the fields produced by \textsc{FINS} are not only geometrically accurate at the surface but also suitable for real–time control tasks requiring gradient and iso–surface information.

\section{Conclusions}
We propose Fast Image-to-Neural Surface (FINS), a framework that reconstructs a high-fidelity SDF field within a few seconds, given either a \emph{single} image input. Our framework combines a multi-resolution hash grid encoder with light-weight geometry head and color head to accelerate convergence. we then leverage a hybrid optimization approach, where an approximate second-order kronecker-factorized optimizer is used to speed up convergence and stability further. We empirically validate our method against a suite of methods against a suite of strong baselines on DTU and BlendedMVS, showing that \textsc{FINS} reaches competitive, and often superior, reconstruction quality while reducing both supervision (down to a single image) and wall-clock optimization time to $\sim$10s on consumer hardware.

\bibliographystyle{IEEEtran}
\bibliography{references}

\end{document}